\documentclass[conference]{IEEEtran}
\IEEEoverridecommandlockouts

\usepackage[T1]{fontenc}

\usepackage[utf8]{inputenc}

\usepackage{graphicx}

\usepackage{xspace}
\usepackage{booktabs}
\usepackage{makecell}
\usepackage{tabularx}
\usepackage[table]{xcolor}
\usepackage{url}
\usepackage[inline]{enumitem}
\usepackage{threeparttable}

\makeatletter
\providecommand{\@LN@col}[1]{}
\providecommand{\@LN}[2]{}
\makeatother

\newcommand{\fours}{\texttt{4S}\xspace}
\newcommand{\threesonel}{\texttt{3S+1L}\xspace}
\newcommand{\twostwol}{\texttt{2S+2L}\xspace}
\newcommand{\fourl}{\texttt{4L}\xspace}
\newcommand{\roundrobin}{\texttt{RoundRobin}\xspace}
\newcommand{\serviceclass}{\texttt{ServiceClass}\xspace}
\newcommand{\generictier}{\texttt{GenericTier}\xspace}
\newcommand{\qualityriskgate}{\texttt{QualityRiskGate}\xspace}
\newcommand{\bottleneckawaretier}{\texttt{BottleneckAwareTier}\xspace}
\newcommand{\ttftbudgetgate}{\texttt{TTFTBudgetGate}\xspace}

\begin{document}

\title{Capacity Planning for Resource-Constrained Clinical LLM Serving}

\author{\IEEEauthorblockN{Yi Tang}
\IEEEauthorblockA{\textit{Population Health Sciences} \\
\textit{Weill Cornell Medicine}\\
New York, USA \\
yit4005@med.cornell.edu}
\and
\IEEEauthorblockN{Geoffrey Martin}
\IEEEauthorblockA{\textit{Systems Engineering} \\
\textit{Cornell University}\\
Ithaca, USA \\
ghm4002@med.cornell.edu}
\and
\IEEEauthorblockN{Yiye Zhang}
\IEEEauthorblockA{\textit{Population Health Sciences} \\
\textit{Weill Cornell Medicine}\\
New York, USA \\
yiz2014@med.cornell.edu}
\and
\IEEEauthorblockN{Yifan Peng}
\IEEEauthorblockA{\textit{Population Health Sciences} \\
\textit{Weill Cornell Medicine}\\
New York, USA \\
yip4002@med.cornell.edu}
}

\maketitle
\begingroup
\renewcommand{\thefootnote}{}
\footnotetext{\copyright~2026 IEEE. Personal use of this material is
permitted. Permission from IEEE must be obtained for all other uses,
in any current or future media, including reprinting/republishing
this material for advertising or promotional purposes, creating new
collective works, for resale or redistribution to servers or lists,
or reuse of any copyrighted component of this work in other works.}
\endgroup
\begin{abstract}
Deploying large language models (LLMs) under a fixed accelerator budget requires jointly deciding how computational capacity is allocated across model sizes and how incoming requests are routed across that capacity. Although prior work has extensively studied LLM routing and serving efficiency, the relative importance of fleet composition and routing under resource constraints remains less understood. We investigate this question using a controlled four-GPU serving environment with heterogeneous clinical NLP workloads spanning short factual verification, biomedical semantic matching, and long-context clinical summarization. We systematically vary fleet composition, routing policy, and offered load while holding the workload trace and hardware budget fixed. Across 24 configurations, descriptive main-effect shares attribute more variation to fleet composition than to the evaluated routing policies in time to first token, latency, and long-context coverage. All-small fleets provide the best responsiveness and throughput but substantially lower long-context coverage, whereas larger-model capacity improves coverage at the cost of serving efficiency. Heterogeneous fleets provide intermediate operating points but can introduce strong-model bottlenecks when large-model capacity is scarce. In a focused evaluation of this bottleneck, capacity-aware routing reduces mean latency and increases throughput while incurring a small reduction in long-context coverage. Within this measured regime, the results support prioritizing capacity allocation, with routing managing demand within the provisioned capacity.
\end{abstract}

\begin{IEEEkeywords}
clinical LLM serving, healthcare AI infrastructure, mixed-model GPU fleets, capacity planning, inference routing, resource-constrained deployment
\end{IEEEkeywords}

\section{Introduction}

Large language models (LLMs) increasingly support heterogeneous clinical NLP workloads, ranging from short interactive prompts and semantic-matching requests to long-context generation.  These workloads impose substantially different computational demands and may also require different levels of model capability. In a resource-constrained deployment, the question is not simply which model should answer a given request, but \textbf{how to allocate limited computational capacity across models of different sizes}. This allocation determines which requests experience contention, how much capacity is available for quality-sensitive workloads, and ultimately the trade-off between inference efficiency and task performance.

This problem is particularly important for clinical applications, where organizations may deploy LLMs on dedicated or self-hosted infrastructure to meet privacy, governance, or operational requirements \cite{rauf2026practice,wiest2025oncology}. Unlike elastic API-based services, such deployments often operate within a fixed accelerator budget. A service may need to accommodate interactive requests while simultaneously processing computationally expensive long-context workloads. Allocating more capacity to smaller models can improve responsiveness and throughput, but may limit access to stronger models for requests that benefit from greater model capacity. Conversely, provisioning more large-model instances can improve coverage of demanding tasks while increasing latency and reducing the efficiency of requests that smaller models could serve. The resulting trade-off is fundamentally a \textbf{capacity-allocation problem}.

\begin{table}[t]
\caption{Experimental factors for mixed-workload inference under a fixed resource envelope. Fleet composition, routing policy, and load are varied independently while the workload trace remains fixed.}
\label{tab:benchmark-design}
\centering
\rowcolors{2}{white}{gray!10}
\begin{tabularx}{\linewidth}{l>{\raggedright\arraybackslash}X>{\raggedright\arraybackslash}X}
\toprule
Component & Realization & Role in evaluation \\
\midrule
Workload & One fixed trace spanning three task families and request-cost profiles & Creates simultaneous demand for fast and strong tiers \\
Arrival & Nominal and overload regimes & When queueing and tail latency emerge \\
Fleet & \fours, \threesonel, \twostwol, \fourl & How a fixed GPU budget should be divided \\
Control & Three main routers plus focused gates & What routing adds after provisioning \\
Outcomes & Serving, GPU, and quality metrics & Which configurations lie on the operational frontier \\
\bottomrule
\end{tabularx}
\end{table}

Prior work has extensively investigated improving LLM inference efficiency through memory management, scheduling, prefill/decode optimization, and multi-model serving \cite{zhong2024distserve,agrawal2024sarathi,patel2024splitwise,yao2024scalellm, muxserve2024,stripelis2024tensoropera,shah2025selectthenroute,aibrix2025}. Another body of work focuses on routing or cascading requests across models with different capabilities \cite{pichlmeier2024expertrouter}. These approaches demonstrate that intelligent request assignment can improve efficiency or quality under a given serving configuration. However, they leave an important question less explored: \textbf{how much does the underlying allocation of model capacity matter relative to the routing strategy}? This distinction matters because routing can redistribute requests among available replicas, but it cannot create additional capacity for an under-provisioned model tier. In particular, a router that directs quality-sensitive requests to a strong model may improve task coverage while simultaneously concentrating demand on a scarce strong-model replica and increasing queueing and latency.

We investigate this interaction between \textbf{capacity allocation and request routing} in a fixed-resource clinical LLM serving setting. We consider a four-GPU deployment using small and large instruction-tuned models from the same family, and we evaluate four capacity allocations: \fours, \threesonel, \twostwol, and \fourl. We expose each allocation to the same heterogeneous workload containing short factual requests, biomedical semantic matching, and long-context clinical summarization. The mixed workload draws on PubHealth \cite{kotonya2020pubhealth}, BioSSES \cite{sogancioglu2017biosses}, and MIMIC-BHC \cite{aali2025mimicivextbhc} to vary request length, serving cost, and measured quality needs. We then compare routing strategies under both nominal and overloaded demand.

Concretely, we make three contributions:
\begin{enumerate*}[nosep, label=(\roman*)]
\item A reproducible clinical mixed-workload study (Table~\ref{tab:benchmark-design}) that factorizes fleet composition, routing policy, and arrival regime while jointly measuring serving, GPU, and task-quality outcomes;
\item An empirical provisioning result showing that allocation across small and large models has a greater effect on time to first token (TTFT), latency, throughput, and long-context coverage than swapping among the evaluated routing policies; and
\item An operational decision analysis showing where heterogeneous fleets balance responsiveness and coverage, and how capacity-aware gating can refine a fixed fleet when large-model capacity is scarce.
\end{enumerate*}

The contribution is an empirical comparison of provisioning and routing in a controlled clinical workload, rather than a new general provisioning algorithm. A motivating scenario is a dedicated service handling short requests alongside hospital-course summarization: assigning summarization to a stronger model is useful only if that tier has sufficient capacity. This scenario motivates the workload design; we do not evaluate hospital traffic or establish workflow-specific latency requirements. The system's mechanisms may also apply to other mixed workloads, and we do not claim that they are unique to healthcare.

\section{Related work}

\begin{figure*}[t]
  \centering
  \includegraphics[width=\textwidth]{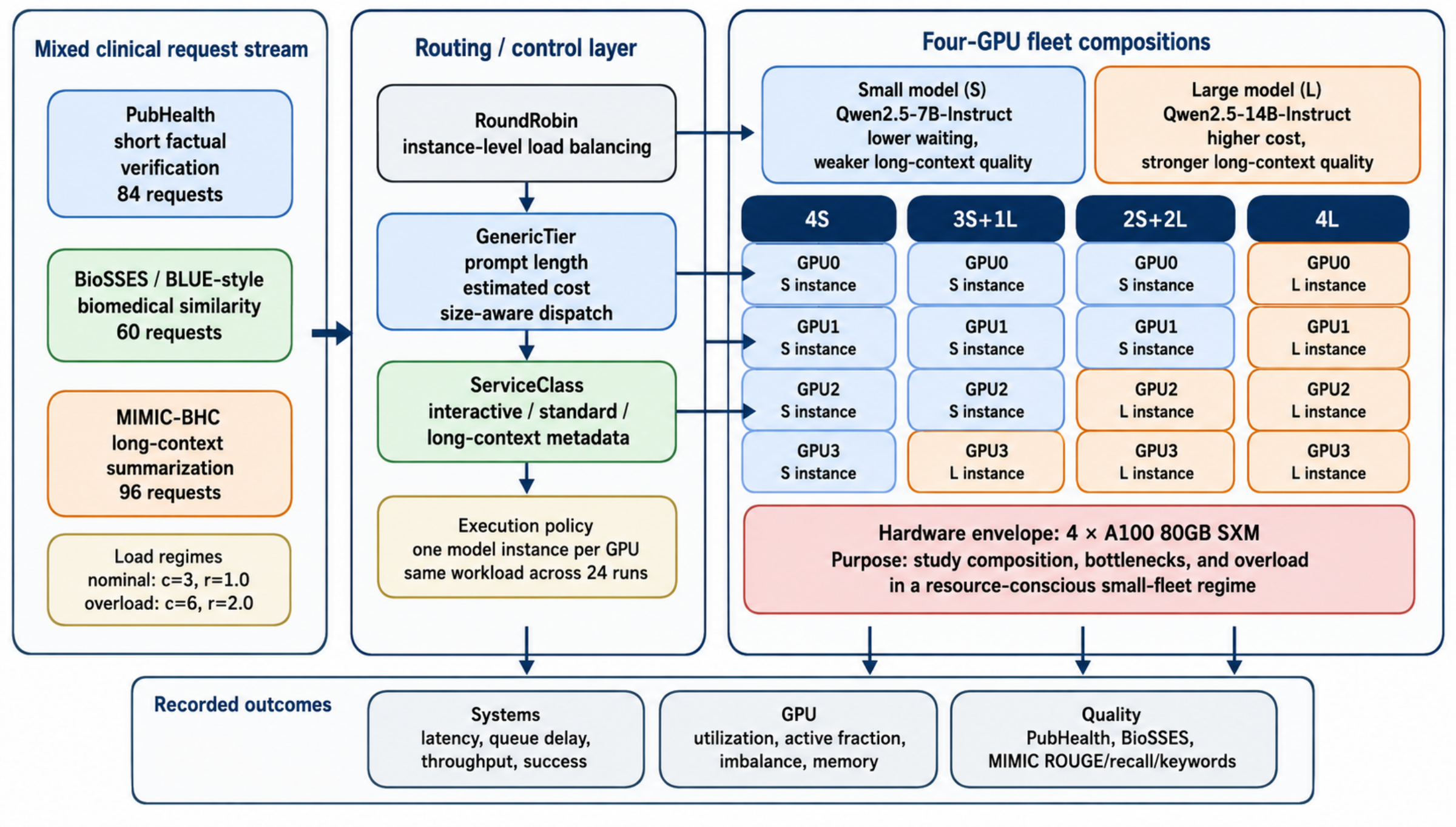}
  \caption{System diagram for the four-GPU study. The displayed 84/60/96 request counts are nominal counts implied by the intended 0.35/0.25/0.40 mixture, not the realized trace; Table~\ref{tab:data} reports the actual 92/59/89 counts. Routing boxes denote alternative policies, not a sequential cascade. Model-quality labels refer to benchmark coverage proxies, not clinical validation.}
  \label{fig:system-diagram}
\end{figure*}

\subsection{Resource allocation for heterogeneous workloads}
The broader systems literature has long recognized that heterogeneous workloads require explicit consideration of both workload characteristics and resource allocation.
CH3 constructs a mixed transactional and analytical workload to expose behavior hidden by single-purpose database tests \cite{mahin2022ch3}. ReAssigner organizes heterogeneous cloud requests around resource roles \cite{cui2022reassigner}, while dynamic workflow provisioning uses monitored resource demand to reduce over-allocation while protecting throughput \cite{bader2023memory}. Foundation-model data systems similarly emphasize end-to-end operational constraints between storage, data access, and accelerators \cite{kienzler2023tensorbank}.

Our setting shares this emphasis on heterogeneous demand but adds another dimension. Resource allocation affects not only serving efficiency but also the model capacity available to different tasks. We use a mixed clinical workload to expose this interaction and distinguish effects arising from fleet composition from those arising from routing and arrival pressure.

\subsection{LLM serving and inference efficiency}
Researchers have studied efficient LLM serving extensively from the perspectives of memory management, scheduling, and inference-pipeline optimization.
PagedAttention improves memory efficiency by managing the key-value cache through paging, enabling higher serving capacity under constrained GPU memory \cite{kwon2023pagedattention}. DistServe separates prefill and decoding to improve the throughput--latency trade-off \cite{zhong2024distserve}, while Sarathi-Serve introduces chunked prefill to better balance throughput and latency \cite{agrawal2024sarathi}. Splitwise further explores phase splitting to improve generative inference efficiency \cite{patel2024splitwise}. Other systems, including ScaleLLM, optimize end-to-end serving efficiency under resource constraints \cite{yao2024scalellm}.

Our work complements this line of research by examining how to divide a fixed accelerator budget across models of different sizes. This distinction matters because the number of replicas allocated to each model tier determines the capacity available to routing policies and can create asymmetric bottlenecks.

\subsection{Multi-model routing}

Another line of work studies serving multiple models and routing requests across heterogeneous model capacities.
Multi-model serving systems such as MuxServe jointly consider spatial and temporal multiplexing to improve resource utilization across multiple LLMs \cite{muxserve2024}. TensorOpera Router investigates routing requests across models to improve inference efficiency \cite{stripelis2024tensoropera}, while subsequent work explores taxonomy-guided routing \cite{shah2025selectthenroute} and unified formulations of routing and cascading \cite{dekoninck2024routing}. RouteLLM learns routing policies from preference data to select among models with different capabilities \cite{ong2024routellm}. Expert Router further characterizes the performance of modular routing configurations across different LLM variants and concurrency levels \cite{pichlmeier2024expertrouter}.

Our research instead focuses on the preceding provisioning decision. Our experiments explicitly separate fleet composition from routing policy, allowing us to quantify their relative effects and to examine routing after the underlying capacity allocation has been fixed.

\subsection{Clinical LLM evaluation and deployment}

Clinical NLP provides a particularly relevant setting for studying these systems questions because clinical workloads vary substantially in input length, computational requirements, and latency tolerance. We use PubHealth for short factual claim verification \cite{kotonya2020pubhealth}, BioSSES for biomedical semantic matching \cite{peng2019blue,sogancioglu2017biosses}, and MIMIC-BHC for long-context hospital-course summarization \cite{aali2025mimicivextbhc}. These datasets offer complementary workload characteristics rather than a unified clinical evaluation benchmark.

Prior work has also examined LLMs in clinical deployment settings. Medical summarization has been integrated into hospital workflows \cite{rauf2026practice}, and local LLM-based information extraction has been explored for privacy-preserving clinical applications \cite{wiest2025oncology}. At the same time, clinical evaluation studies emphasize that automatic benchmark metrics do not fully capture medical usefulness or safety \cite{zhang2025llmevalmed}.

Our work differs from prior clinical LLM studies in that clinical tasks serve as heterogeneous workloads for investigating a systems-level research question. We do not seek to identify the clinically best model. Instead, we examine how model capacity allocation and routing jointly determine the operational behavior of a resource-constrained clinical LLM service.

\section{Methods}

We study LLM serving under a fixed accelerator budget, with multiple model sizes serving a heterogeneous stream of clinical NLP requests (Figure~\ref{fig:system-diagram}).
We therefore separate the serving configuration into three factors: (1) fleet composition; (2) routing policy; and (3) workload.

\subsection{Fleet composition}

We model a clinical inference tier with a fixed accelerator budget rather than an elastic public API. We use a vLLM-based multi-instance serving stack with one model instance per GPU \cite{kwon2023pagedattention}. The main study uses a four-GPU pod. All four accelerators are identical A100s; heterogeneity comes from allocating the serving slots to different model sizes, not from mixing GPU types. Here, \emph{resource-constrained} means a fixed number of accelerator slots rather than low-end hardware. This scale exposes contention and composition effects within a fixed local-deployment scenario; it is not a measured estimate of a hospital's required capacity.

The main experimental variable is fleet composition.
\begin{enumerate*}[nosep, label=(\arabic*)]
    \item \fours: four small-model instances.
    \item \threesonel: three small-model instances and one large-model instance.
    \item \twostwol: two small-model instances and two large-model instances.
    \item \fourl: four large-model instances.
\end{enumerate*}

We do not claim that these models are clinically optimal. Using two instruction-tuned sizes from the same Qwen2.5 family \cite{yang2024qwen25} provides a controlled comparison of replica allocation while reducing tokenizer and training-family differences; it also limits generalizability beyond this model family.

The small model is Qwen2.5-7B-Instruct, and the large model is Qwen2.5-14B-Instruct.
Both the original and reliability execution blocks used four NVIDIA A100-SXM4-80GB accelerators with one model server per GPU. The reliability block ran on a second, nominally matched four-A100 pod; its accelerator inventory, model revisions, runtime environment, and run provenance were archived with the measurements.

\subsection{Routing policies}

Table~\ref{tab:routing-rules} summarizes the implemented routing rules. Requests assigned to a tier are distributed round-robin among that tier's instances. If no instance of the requested tier exists, all fleet instances form the fallback pool. The policies do not use a learned controller or live queue-length feedback.

\begin{table*}[t]
\caption{Routing rules used in the main study and focused follow-up. ``Risk'' is the deterministic profile score produced from task and request metadata; token thresholds refer to estimated prompt or maximum output length.}
\label{tab:routing-rules}
\centering
\rowcolors{2}{white}{gray!10}
\begin{tabularx}{\textwidth}{ll>{\raggedright\arraybackslash}X}
\toprule
Policy & Primary signal & Implemented rule \\
\midrule
\roundrobin & Instance index & Cycle over all instances without task or model-tier semantics. \\
\generictier & Length & Select the strong tier when prompt length $\geq1700$ tokens or maximum output length $\geq224$ tokens; otherwise select the fast tier. \\
\serviceclass & Metadata and risk & Select the strong tier for standard-generation, long-context, background, or summarization requests, or when risk $\geq0.60$. \\
\qualityriskgate & Risk and length & Use risk thresholds 0.62/0.42, length thresholds 1700/224, and clinical-keyword density 0.08 to gate strong-tier escalation. \\
\bottleneckawaretier & Fleet composition & Treat strong capacity as scarce when its fleet fraction is $\leq0.34$; then use stricter risk thresholds 0.72/0.52 together with 1700/224 length thresholds. \\
\ttftbudgetgate & Risk and prefill budget & Use risk thresholds 0.72/0.50, prefill budgets 1400/2200, output threshold 224, and strong-fleet scarcity 0.34 to protect first-token responsiveness. \\
\bottomrule
\end{tabularx}
\end{table*}

We evaluate three simple, interpretable routing policies across fleet compositions.
\begin{enumerate*}[nosep, label=(\roman*)]
    \item \roundrobin: instance-level load balancing without workload semantics.
    \item \generictier: size-aware routing using generic cost signals such as prompt length and estimated inference cost.
    \item \serviceclass: size-aware routing using workload metadata, such as interactive, standard-generation, and long-context service classes.
\end{enumerate*}
These policies use fixed, interpretable gateway rules without training a specialized router. Their metadata-based risk scores are serving heuristics, not validated assessments of clinical risk.

To test whether routing still matters after fleet composition has been fixed, we run an eight-configuration follow-up on the bottleneck-prone \threesonel fleet. We use \generictier as the baseline and add three capacity-aware policies aligned with standard serving objectives:
\begin{enumerate*}[nosep, label=(\roman*)]
\item \qualityriskgate: escalates requests estimated to have higher quality risk.
\item \bottleneckawaretier: tightens escalation when the large-model fraction of the fleet is scarce.
\item \ttftbudgetgate: limits escalation when it would threaten a TTFT budget.
\end{enumerate*}
These policies are not introduced as universal routers. They test whether a fixed mixed-model fleet can improve observed latency and throughput by protecting scarce strong-model capacity.

\subsection{Workload}

Each run consists of 240 requests sampled from a mixed clinical workload.
The workload combines three benchmark-derived clinical NLP task families (Table~\ref{tab:data}):
\begin{enumerate*}[nosep, label=(\roman*)]
\item \textbf{PubHealth} claim verification \cite{kotonya2020pubhealth}, which models short factual requests;
\item \textbf{BioSSES} biomedical similarity \cite{peng2019blue,sogancioglu2017biosses}, which models short semantic matching requests; and
\item \textbf{MIMIC-BHC} long-context clinical summarization \cite{aali2025mimicivextbhc}, which models expensive longitudinal generation requests.
\end{enumerate*}
Fixed prompts require a label-only PubHealth response, a 0--5 BioSSES similarity score, and a hospital-course summary for MIMIC-BHC; the released workload adapter contains the exact templates.

\begin{table}[t]
\caption{Workload composition.
}
\label{tab:data}
\centering
\rowcolors{2}{white}{gray!10}
\begin{tabularx}{\linewidth}{ll>{\raggedright\arraybackslash}X>{\raggedright\arraybackslash}X}
\toprule
Dataset family & Requests & Prompt pattern & Serving role \\
\midrule
PubHealth & 92 & Short public-health claim verification with label-style output & Fast, latency-sensitive factual requests \\
BioSSES & 59 & Short sentence-pair similarity & Moderate semantic reasoning \\
MIMIC-BHC & 89 & Long-context EHR summarization & Expensive, quality-sensitive generation \\
\bottomrule
\end{tabularx}
\end{table}

The intended sampling weights are 0.35 for PubHealth, 0.25 for BioSSES, and 0.40 for MIMIC-BHC. A deterministic generator with seed 42 produced the realized 92/59/89 request counts in Table~\ref{tab:data} and the same generated request order for matched configurations. The fixed trace interleaves bursts of two long requests with four short requests. For PubHealth and BioSSES, synthetic prefix reuse is enabled with probability 0.6 across six prefix groups. Decoding was deterministic (temperature 0), and the benchmark driver's batch-size parameter was 1.

We evaluate two operating regimes:
\begin{enumerate*}[nosep, label=(\roman*)]
    \item \textbf{Nominal load} with concurrency 3 and request rate 1.0 requests/s, and
    \item \textbf{Overload} with concurrency 6 and request rate 2.0 requests/s.
\end{enumerate*}

\subsection{Experimental protocol}

The complete 24-configuration matrix and eight-configuration follow-up were initially executed once per cell (\texttt{trials: 1}). We subsequently executed 42 fresh process-level reliability runs on the second pod while holding the trace, request order, decoding, routing logic, and load controls fixed. To test the paper's two central claims directly, this block contains four repeats of all eight \generictier fleet-by-load operating points and five repeats of the two \threesonel \bottleneckawaretier points. We retain the original matrix as the complete factorized study and use the repeated block to quantify process-level uncertainty for these ten operating points. Repeating the same trace measures execution variability, not uncertainty over different patients, request mixtures, or arrival traces.

Canonical experiment manifests and YAML configurations, together with the workload generator, routing implementations, and execution scripts, are available at \url{https://anonymous.4open.science/r/clinical-llm-serving-configs-A35F}. The release excludes private dataset contents and machine-specific logs. The anonymized artifact URL is retained as the access endpoint; the authors are identified in this manuscript.

\begin{figure*}[t]
  \centering
  \includegraphics[width=\linewidth]{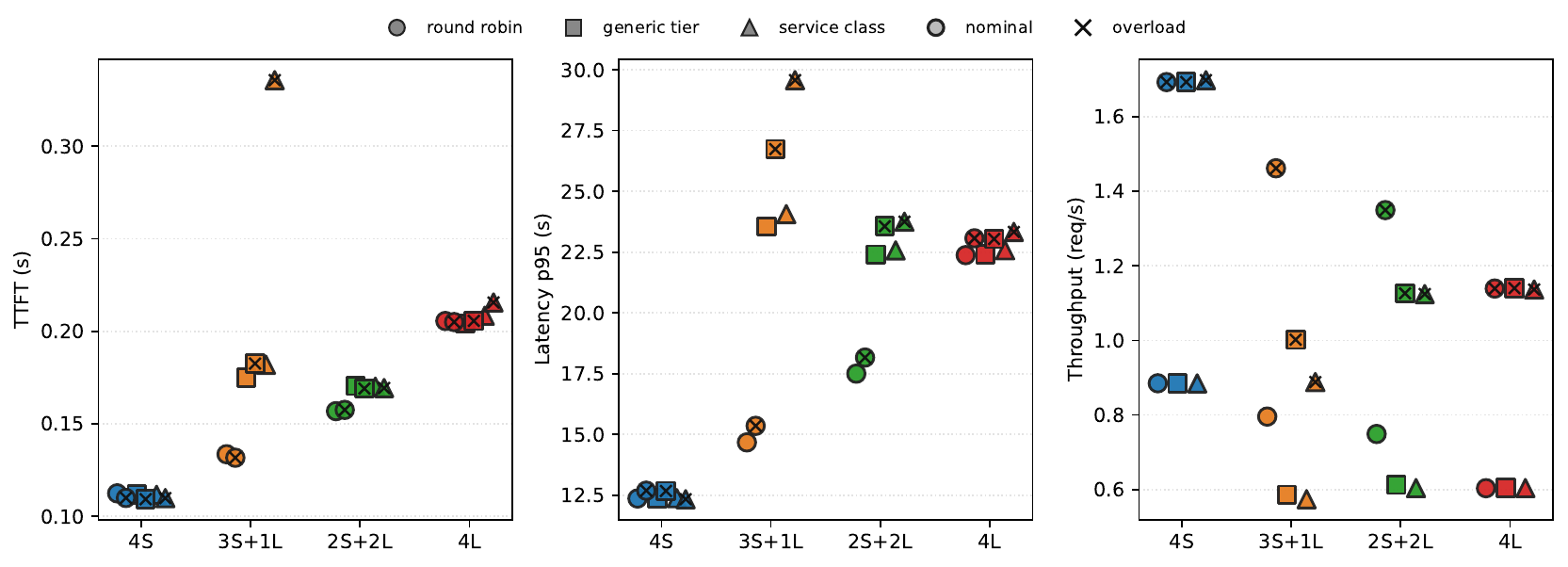}
  \caption{Primary serving metrics for the four-GPU fleet study. Across TTFT, p95 latency, and throughput, fleet composition shifts the operating regime much more than routing-policy swaps do. Marker shape denotes policy, and overload points are marked with an x.}
  \label{fig:fleet-systems}
\end{figure*}

\subsection{Evaluation metrics}

We evaluate the system along three dimensions: serving performance, resource utilization, and task performance.

\subsubsection{Serving performance}
We report TTFT, mean request service latency, p95 service latency, throughput, and success rate, following the standard serving-oriented metric vocabulary used in contemporary benchmarking and deployment tooling \cite{nvidia2026nimmetrics,anyscale2026metrics,vllm2026servebench}. Service latency is measured from dispatch to completion; queue delay separately measures the interval from planned arrival to dispatch. Because the two load regimes jointly change request rate and worker concurrency, queue delay is compared among configurations within a regime rather than interpreted as monotonic across regimes.
We also report the mean and p95 queue delay as diagnostic metrics that distinguish slow generation from slow admission to scarce capacity.
Because all runs completed successfully and we do not impose a latency service-level objective, we report throughput rather than SLO-qualified goodput. Service-class labels do not specify validated clinical deadlines; we do not evaluate emergency-workflow SLA violations.

\subsubsection{Resource utilization}

We monitor mean utilization, active fraction, memory occupancy, and cross-GPU imbalance to characterize whether a heterogeneous fleet uses all capacity or whether a single large-model server becomes the dominant bottleneck.

\subsubsection{Benchmark quality}
For PubHealth, we report accuracy and macro-F1.
For BioSSES, we report Pearson correlation, with Mean Absolute Error (MAE) and Root Mean Square Error (RMSE) used as auxiliary diagnostics.
For MIMIC-BHC, we report ROUGE-L, token recall, token precision, and clinical keyword recall.
The main text emphasizes MIMIC token recall because it directly measures coverage of long-context reference content.
These automatic metrics are task-performance proxies rather than measures of clinical factuality, safety, or usefulness. Token recall can favor longer outputs and does not establish whether the added content is correct. Quality metrics are reported without confidence intervals, so small differences are descriptive. With 92 PubHealth requests, one changed prediction corresponds to about 1.09 percentage points in accuracy.

Finally, to compare fleet and routing effects in the balanced main matrix, we compute each factor's descriptive main-effect share of total sum of squares across the 24 point estimates. These shares are not inferential significance tests, and the unassigned variation includes factor interactions.

\section{Results and Discussion}

\subsection{Fleet composition dominates serving behavior}

Within the evaluated model, hardware, and workload regime, fleet composition produced larger observed changes than the three main routing policies. Figure~\ref{fig:fleet-systems} summarizes serving metrics, Table~\ref{tab:fleet-main} highlights operating points, and Table~\ref{tab:full-matrix-summary} preserves the full matrix for reproducibility. Descriptive main-effect shares for fleet versus policy are 52.8\% versus 8.8\% for TTFT, 65.3\% versus 13.2\% for mean latency, 65.8\% versus 13.6\% for p95 latency, 22.6\% versus 3.2\% for throughput, and 75.3\% versus 11.8\% for MIMIC token recall. Load, rather than fleet, is the largest main effect for throughput.

\subsubsection{All-small fleets are fastest}
The \fours fleet consistently delivers the strongest responsiveness. Under nominal load, TTFT is approximately 0.112s and mean latency is 2.63--2.64s. Under overload, TTFT remains near 0.109--0.110s while throughput rises to about 1.69--1.70 requests/s. However, MIMIC token recall remains near 0.19, indicating weaker long-context coverage.

\subsubsection{All-large fleets are coverage-heavy but slower}
The \fourl fleet provides stronger long-context coverage than \fours but worse serving behavior. Under nominal load, TTFT is approximately 0.204--0.208s, mean latency is near 4.79s, and p95 latency exceeds 22s. Under overload, latency remains around 4.94s, and p95 latency stays near 23s, with GPU utilization above 85\%. \fourl is therefore best interpreted as a coverage-heavy reference point rather than a default deployment choice.

\subsubsection{A single large instance can become a bottleneck}
The \threesonel fleet recovers much of the long-context coverage frontier, but the single large-model instance can also become a scarce path. Under nominal load, \threesonel with \generictier achieves the highest MIMIC token recall in the main matrix (0.2700), but TTFT rises to 0.175s, mean latency to 4.92s, and p95 latency to 23.55s. Under overload, token recall remains high at 0.2664, but p95 latency increases to 26.74s. In the nominal request log, \generictier directs 86 requests to the single large replica and 154 across three small replicas; mean queue delay is 54.35s on the large path versus 41.96s across the small paths. This concentration suggests a strong-path bottleneck rather than fleet-wide saturation.

\subsubsection{Balanced heterogeneity improves robustness}

The \twostwol fleet is a candidate when responsiveness and measured coverage both matter. Under overload, \twostwol with \generictier reaches mean latency 5.007s and throughput 1.126 requests/s, compared with 5.681s and 1.002 requests/s for \threesonel. Its observed MIMIC recall is 0.263 versus 0.266; this small coverage difference is not a ranking factor.

A single strong-model instance may recover most measured coverage gains, but it can also become a bottleneck. If future clinical agents increase the share of long-context calls, capacity pressure is likely to concentrate on stronger models. Consequently, fleet-design decisions can substantially affect observed service responsiveness.
\begin{table}[t]
  \caption{Representative operating points from the GPU fleet matrix. }
  \label{tab:fleet-main}
  \centering
  \setlength{\tabcolsep}{4pt}
  \begin{threeparttable}
  \begin{tabular}{lllccccc}
  \toprule
  Fleet & Load & Policy & \makecell[c]{TTFT\\(s)} & \makecell[c]{Latency\\(s)} & \makecell[c]{p95\\(s)} & \makecell[c]{Throughput\\(requests/s)} & \makecell[c]{MIMIC\\recall} \\
  \midrule
  \fours & N & RR & 0.112 & 2.640 & 12.37 & 0.885 & 0.190 \\
  \fours & O & RR & 0.110 & 2.693 & 12.69 & 1.693 & 0.192 \\
  \rowcolor{gray!10}
  \threesonel & N & GT & 0.175 & 4.918 & 23.55 & 0.586 & 0.270 \\
  \rowcolor{gray!10}
  \threesonel & O & GT & 0.183 & 5.681 & 26.74 & 1.002 & 0.266 \\
  \twostwol & N & GT & 0.170 & 4.685 & 22.40 & 0.613 & 0.264 \\
  \twostwol & O & GT & 0.169 & 5.007 & 23.56 & 1.126 & 0.263 \\
  \rowcolor{gray!10}
  \fourl & N & RR & 0.206 & 4.795 & 22.38 & 0.604 & 0.265 \\
  \rowcolor{gray!10}
  \fourl & O & RR & 0.205 & 4.943 & 23.07 & 1.139 & 0.264 \\
  \bottomrule
  \end{tabular}
  \begin{tablenotes}[para, flushleft]
  \footnotesize
  \item[] Abbreviations: N, nominal; O, overload; RR, \roundrobin; GT, \generictier.
  \end{tablenotes}
  \end{threeparttable}
\end{table}

\subsection{Mixed-model fleets define the coverage--latency frontier}

\begin{figure}[t]
  \centering
  \includegraphics[width=\columnwidth]{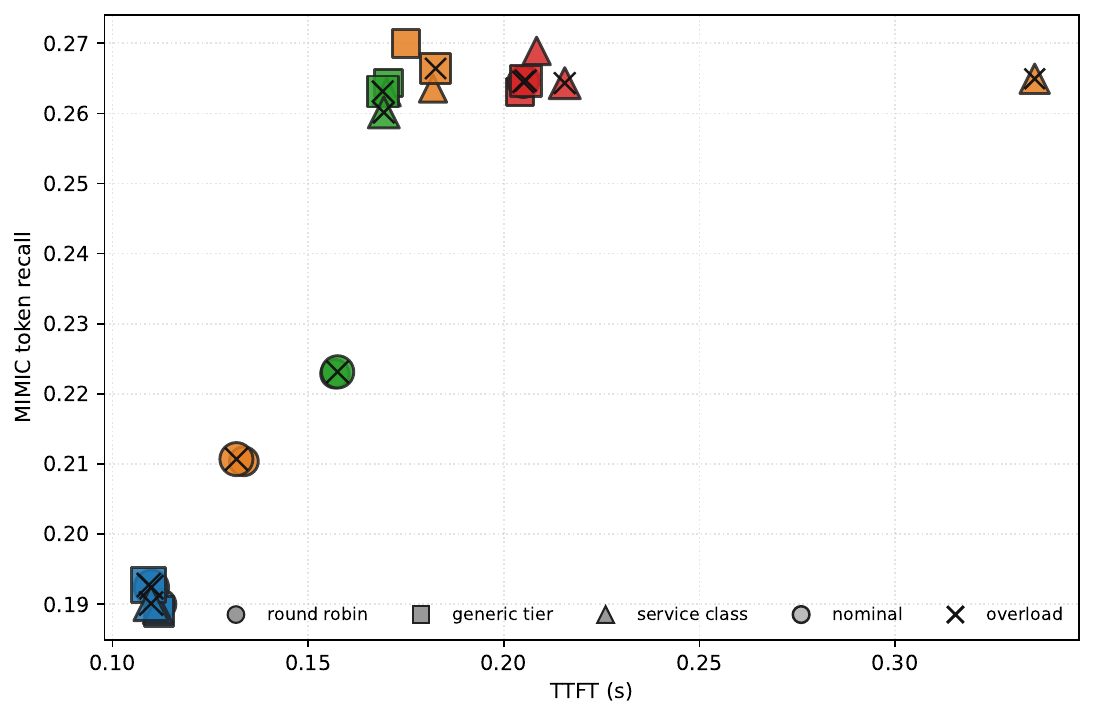}
  \caption{Observed TTFT--coverage trade-offs in the original single-run matrix. The vertical axis is MIMIC token recall, not clinical factuality or safety; small coverage gaps have no reported confidence intervals.}
  \label{fig:tradeoff-frontier}
\end{figure}

Table~\ref{tab:fleet-main} reports representative operating points from the main matrix, and Figure~\ref{fig:tradeoff-frontier} visualizes it in TTFT--coverage space. The frontier is not captured by a single best configuration because deployment objectives differ in acceptable latency and measured coverage thresholds.

The main operational distinction is between the lower measured coverage of \fours and the higher coverage of configurations that route long requests to stronger models. The much smaller coverage differences among strong-tier configurations do not establish a unique best fleet. Figure~\ref{fig:tradeoff-frontier} shows the original within-block trade-offs; it is not a confidence-bounded or clinically validated frontier.

For clinical deployments, the main question is not only \emph{which model should answer this request?} but also \emph{how many instances of each model size should exist in the serving fleet?}

\subsection{Routing is secondary but informative in bottlenecked fleets}

Routing policy had relatively small effects within homogeneous fleets. Within \fours and \fourl, \roundrobin, \generictier, and \serviceclass differ by only a few milliseconds in TTFT, a few hundredths of a second in mean latency, and very little in task metrics. In the measured regime, routing therefore does not compensate for the fleet's underlying capacity profile.

The effects were somewhat larger in heterogeneous fleets, but remained secondary to fleet composition. \serviceclass occasionally helps short-task quality, but it does not consistently outperform \generictier on long-context quality or serving efficiency. For example, in \threesonel under nominal load, \generictier achieves MIMIC token recall of 0.270, while \serviceclass reaches 0.264 and worsens TTFT from 0.175s to 0.182s.
In \threesonel under overload, \serviceclass raises TTFT to 0.336s, compared with 0.183s for \generictier.

These point estimates are consistent with metadata-based escalation amplifying contention on a single large-model instance. This comparison concerns the evaluated fixed rules; it does not establish the relative performance of learned or live queue-aware routers.

Because the complete matrix was initially executed once per cell, small gaps involving the unrepeated \roundrobin and \serviceclass policies remain descriptive. The reliability block provides repeated estimates for the selected \generictier fleet points and the \bottleneckawaretier comparison, while the repeated serving measurements support the selected fleet contrasts. Subsecond TTFT differences alone do not establish clinical benefit; the multi-second service-latency and queue-delay differences motivate the capacity analysis.

\subsection{Capacity-aware routing improves serving efficiency}

We next asked whether routing could improve performance after fleet composition had been fixed. We restrict the reported follow-up to \threesonel because it directly tests the policy question in the fleet where the main matrix identifies scarce strong-model capacity. This avoids treating measurements collected in a separate \twostwol execution block as interchangeable with the main-matrix \twostwol operating points.

Table~\ref{tab:followup-3s1l} reports the original focused follow-up point estimates. In the bottleneck-prone \threesonel fleet, \bottleneckawaretier has better observed serving metrics than \generictier. Under nominal load, it reduces TTFT from 0.173s to 0.165s, mean latency from 5.00s to 4.43s, p95 latency from 23.77s to 20.89s, increases throughput from 0.578 to 0.655 requests/s, and reduces queue delay from 47.77s to 34.06s. Under overload, the same pattern holds. The reliability analysis below tests this comparison across repeated runs.
Within these point estimates, protecting scarce strong-model capacity is more effective than adding semantic structure alone.
\begin{table}[t]
  \caption{Original single-run \threesonel follow-up using capacity-aware routing variants. }
  \label{tab:followup-3s1l}
  \centering
  \begin{threeparttable}
  \begin{tabular}{llccccc}
  \toprule
  Load & Policy & \makecell[c]{TTFT\\(s)} & \makecell[c]{Latency\\(s)} & \makecell[c]{p95\\(s)} & \makecell[c]{Throughput\\(requests/s)} & \makecell[c]{Queue\\(s)} \\
  \midrule
  N & GT  & 0.173 & 4.995 & 23.77 & 0.578 & 47.77 \\
  N & QRG & 0.170 & 4.981 & 23.74 & 0.579 & 47.50 \\
  N & BAT & 0.165 & 4.429 & 20.89 & 0.655 & 34.06 \\
  N & TBG & 0.157 & 4.512 & 22.86 & 0.634 & 35.77 \\
  \midrule
  O & GT  & 0.181 & 5.661 & 26.74 & 1.005 & 30.60 \\
  O & QRG & 0.175 & 5.635 & 26.47 & 1.010 & 30.40 \\
  O & BAT & 0.162 & 4.920 & 23.80 & 1.172 & 22.00 \\
  O & TBG & 0.165 & 4.983 & 23.94 & 1.101 & 22.69 \\
  \bottomrule
  \end{tabular}
  \begin{tablenotes}[para, flushleft]
  \footnotesize
  \item[] Abbreviations: N, nominal; O, overload; GT, \generictier; QRG, \qualityriskgate; BAT, \bottleneckawaretier; TBG, \ttftbudgetgate.
  \end{tablenotes}
  \end{threeparttable}
\end{table}

\subsection{Reliability and cross-environment robustness}

The repeated runs also quantify the variability of the serving measurements (Table~\ref{tab:reliability-reruns}). Values are the mean $\pm$ 95\% Student-$t$ CI half-width. Across the ten repeated operating points, coefficients of variation are 0.3--2.0\% for TTFT, 0.3--0.9\% for mean latency, 0.5--1.7\% for p95 latency, and 0.3--1.2\% for throughput. Mean queue delay is comparatively more variable, with coefficients of variation from 1.5\% to 5.6\%, but remains substantially more stable than the fleet-level separations.

Across four paired repeats, \bottleneckawaretier reduces \threesonel mean latency by 11.1\% under nominal load and 12.9\% under overload, reduces p95 latency by 11.6\% and 13.7\%, increases throughput by 13.0\% and 17.0\%, and reduces mean queue delay by 28.4\% and 27.5\%, respectively. The paired 95\% confidence intervals for TTFT, mean latency, p95 latency, throughput, and queue delay all exclude zero. Across the same four paired repeats, these serving gains accompany a small decrease in mean MIMIC-BHC token recall: 0.2371 versus 0.2442 under nominal load and 0.2433 versus 0.2516 under overload. Capacity-aware gating is therefore an explicit capacity--coverage trade-off rather than a free improvement.

The repeated serving measurements preserve the original fleet ordering, but absolute MIMIC-BHC overlap scores differ across pods by up to 0.026. For example, \threesonel \generictier token recall is 0.2442 across the fresh runs versus 0.2700 in the original matrix. We therefore do not treat the original point estimate as a fifth repeat or pool quality measurements across execution environments. The original matrix supports a descriptive within-block coverage comparison, while the fresh block quantifies serving uncertainty for selected points. Because the cross-pod coverage shift exceeds several routing-policy differences, we do not infer a reliable ordering from those small quality gaps.

\begin{table*}[t]
\caption{Fresh process-level reliability runs using the fixed 240-request trace.}
\label{tab:reliability-reruns}
\centering
\begin{threeparttable}
\begin{tabular}{lllcccccr}
\toprule
Fleet & Load & Policy & $n$ & TTFT (s) & Latency (s) & p95 (s) & Throughput & Queue (s) \\
\midrule
\fours & N & GT & 4 & 0.119 $\pm$ 0.004 & 2.691 $\pm$ 0.040 & 12.551 $\pm$ 0.349 & 0.881 $\pm$ 0.004 & 2.680 $\pm$ 0.198 \\
\fours & O & GT & 4 & 0.112 $\pm$ 0.003 & 2.755 $\pm$ 0.031 & 12.907 $\pm$ 0.223 & 1.685 $\pm$ 0.007 & 1.091 $\pm$ 0.097 \\
\threesonel & N & GT & 4 & 0.178 $\pm$ 0.002 & 5.011 $\pm$ 0.047 & 23.743 $\pm$ 0.178 & 0.576 $\pm$ 0.005 & 48.523 $\pm$ 1.789 \\
\threesonel & N & BAT & 5 & 0.167 $\pm$ 0.002 & 4.467 $\pm$ 0.043 & 20.985 $\pm$ 0.192 & 0.650 $\pm$ 0.005 & 35.314 $\pm$ 1.667 \\
\threesonel & O & GT & 4 & 0.187 $\pm$ 0.002 & 5.680 $\pm$ 0.083 & 27.475 $\pm$ 0.700 & 0.998 $\pm$ 0.019 & 30.851 $\pm$ 1.586 \\
\threesonel & O & BAT & 5 & 0.167 $\pm$ 0.003 & 4.940 $\pm$ 0.036 & 23.678 $\pm$ 0.235 & 1.169 $\pm$ 0.008 & 22.265 $\pm$ 0.553 \\
\twostwol & N & GT & 4 & 0.174 $\pm$ 0.001 & 4.732 $\pm$ 0.043 & 22.449 $\pm$ 0.325 & 0.608 $\pm$ 0.006 & 40.540 $\pm$ 0.950 \\
\twostwol & O & GT & 4 & 0.172 $\pm$ 0.002 & 5.059 $\pm$ 0.051 & 23.773 $\pm$ 0.459 & 1.116 $\pm$ 0.011 & 22.080 $\pm$ 0.760 \\
\fourl & N & GT & 4 & 0.210 $\pm$ 0.001 & 4.857 $\pm$ 0.036 & 22.747 $\pm$ 0.267 & 0.597 $\pm$ 0.004 & 43.666 $\pm$ 1.223 \\
\fourl & O & GT & 4 & 0.208 $\pm$ 0.001 & 4.993 $\pm$ 0.026 & 23.410 $\pm$ 0.341 & 1.127 $\pm$ 0.007 & 21.075 $\pm$ 0.542 \\
\bottomrule
\end{tabular}
  \begin{tablenotes}[para, flushleft]
  \footnotesize
  \item[] Abbreviations: N, nominal; O, overload; GT, \generictier; BAT, \bottleneckawaretier.
  \end{tablenotes}
  \end{threeparttable}
\end{table*}

\begin{table*}[t]
\caption{Complete 24-configuration four-GPU matrix.}
\label{tab:full-matrix-summary}
\centering
\begin{threeparttable}
\begin{tabular}{lllrcccrccc}
\toprule
Fleet & Load & Policy & \makecell[c]{TTFT\\(s)} & \makecell[c]{Latency\\(s)} & \makecell[c]{p95\\(s)} & \makecell[c]{Throughput\\(requests/s)} & Queue & \makecell[c]{GPU util.\\(\%)} & \makecell[c]{PubHealth\\acc.} & \makecell[c]{MIMIC\\recall} \\
\midrule
\fours & N & RR & 0.112 & 2.640 & 12.37 & 0.885 & 2.412 & 51.68 & 0.6196 & 0.1900 \\
\fours & N & GT & 0.112 & 2.628 & 12.38 & 0.885 & 2.401 & 51.05 & 0.6196 & 0.1889 \\
\fours & N & SC & 0.112 & 2.635 & 12.38 & 0.885 & 2.409 & 51.58 & 0.6196 & 0.1893 \\
\fours & O & RR & 0.110 & 2.693 & 12.69 & 1.693 & 0.972 & 74.51 & 0.6196 & 0.1924 \\
\fours & O & GT & 0.109 & 2.703 & 12.68 & 1.693 & 0.979 & 73.93 & 0.6196 & 0.1927 \\
\fours & O & SC & 0.110 & 2.689 & 12.34 & 1.698 & 0.943 & 74.25 & 0.6196 & 0.1901 \\
\midrule
\threesonel & N & RR & 0.134 & 3.191 & 14.67 & 0.796 & 5.568 & 53.51 & 0.6413 & 0.2104 \\
\threesonel & N & GT & 0.175 & 4.918 & 23.55 & 0.586 & 46.399 & 25.78 & 0.6087 & 0.2700 \\
\threesonel & N & SC & 0.182 & 5.061 & 24.07 & 0.574 & 49.108 & 25.00 & 0.6196 & 0.2635 \\
\threesonel & O & RR & 0.132 & 3.297 & 15.35 & 1.462 & 2.213 & 72.60 & 0.6413 & 0.2107 \\
\threesonel & O & GT & 0.183 & 5.681 & 26.74 & 1.002 & 30.974 & 26.80 & 0.6087 & 0.2664 \\
\threesonel & O & SC & 0.336 & 6.475 & 29.58 & 0.888 & 62.965 & 22.34 & 0.6196 & 0.2649 \\
\midrule
\twostwol & N & RR & 0.157 & 3.673 & 17.50 & 0.749 & 12.659 & 56.78 & 0.6630 & 0.2229 \\
\twostwol & N & GT & 0.170 & 4.685 & 22.40 & 0.613 & 36.951 & 46.37 & 0.6087 & 0.2643 \\
\twostwol & N & SC & 0.170 & 4.784 & 22.57 & 0.604 & 40.917 & 46.31 & 0.6196 & 0.2630 \\
\twostwol & O & RR & 0.157 & 3.869 & 18.16 & 1.349 & 6.350 & 75.97 & 0.6630 & 0.2231 \\
\twostwol & O & GT & 0.169 & 5.007 & 23.56 & 1.126 & 21.342 & 50.08 & 0.6087 & 0.2631 \\
\twostwol & O & SC & 0.169 & 5.054 & 23.75 & 1.124 & 21.955 & 48.96 & 0.6196 & 0.2601 \\
\midrule
\fourl & N & RR & 0.206 & 4.795 & 22.38 & 0.604 & 41.821 & 63.45 & 0.6630 & 0.2649 \\
\fourl & N & GT & 0.204 & 4.789 & 22.42 & 0.605 & 41.542 & 63.29 & 0.6630 & 0.2630 \\
\fourl & N & SC & 0.208 & 4.790 & 22.58 & 0.604 & 41.515 & 63.46 & 0.6630 & 0.2689 \\
\fourl & O & RR & 0.205 & 4.943 & 23.07 & 1.139 & 20.365 & 85.76 & 0.6630 & 0.2645 \\
\fourl & O & GT & 0.206 & 4.944 & 23.04 & 1.140 & 20.378 & 86.38 & 0.6630 & 0.2647 \\
\fourl & O & SC & 0.216 & 4.949 & 23.33 & 1.137 & 20.520 & 85.97 & 0.6630 & 0.2643 \\
\bottomrule
\end{tabular}
  \begin{tablenotes}[para, flushleft]
  \footnotesize
  \item[] Abbreviations: N, nominal; O, overload; RR, \roundrobin; GT, \generictier; SC, \serviceclass.
  \end{tablenotes}
  \end{threeparttable}
\end{table*}

\subsection{Deployment implications}

The study yields four practical starting hypotheses for healthcare teams operating resource-constrained LLM services.
\textbf{First}, fleet composition should be determined before tuning the router: compared with routing refinements, fleet choice produces larger observed differences in TTFT, latency, throughput, and long-context coverage.
\textbf{Second}, homogeneous small-model fleets are strongest for responsiveness-first applications, while measured long-context coverage increases when requests access large-model capacity. However, adding only one large-model instance can shift the limiting factor from coverage to queueing.
\textbf{Third}, balanced mixed-model fleets offer the best observed compromise when both coverage and responsiveness matter. In our setting, \twostwol avoids the weakest long-context coverage of \fours while reducing the bottlenecks observed in \threesonel.
\textbf{Finally}, capacity-aware routing can refine a bottleneck-prone heterogeneous fleet: by tightening strong-tier escalation when large-model capacity is scarce, \bottleneckawaretier improves TTFT, p95 latency, throughput, and queue delay in the \threesonel follow-up.

Table~\ref{tab:operator-guide} converts these results into starting points rather than universal prescriptions. A deployment should first choose a fleet for its dominant service objective, measure the resulting tier pressure, and only then tune escalation.

\begin{table}[t]
\caption{Operator guide derived from the measured four-GPU regime. Exact choices require recalibration for other models and workloads.}
\label{tab:operator-guide}
\centering
\rowcolors{2}{white}{gray!20}
\begin{tabularx}{\linewidth}{l l >{\raggedright\arraybackslash}X}
\toprule
Primary objective & Starting fleet & Operational interpretation \\
\midrule
Responsiveness & \fours & Lowest TTFT and highest throughput; weaker long-context coverage \\
Balanced service & \twostwol & Reduces dependence on one strong server while retaining measured coverage \\
Scarce strong tier & \threesonel & Use capacity-aware escalation to avoid concentrating the queue \\
Coverage-heavy & \fourl & Strong-model capacity throughout, with a slower serving regime \\
\bottomrule
\end{tabularx}
\end{table}

For a clinical service, these results provide capacity-planning hypotheses to test against its own request mix and workflow targets. They do not establish production readiness, emergency response performance, or clinical safety.

\subsection{Scope and transferability}

The absolute operating points in this study are specific to the Qwen2.5-7B/14B service-rate ratio, the four-A100 resource envelope, and the realized request mix. The study instead provides three mechanisms to test in other deployments: composition sets the aggregate service capacity of each tier; asymmetric tier capacity can concentrate queueing on the strong path; and stricter escalation can relieve that contention when quality demand permits. Their boundaries will move with model-speed ratios, context-length distributions, batching, cache behavior, accelerator type, and arrival process. Although motivated by clinical data governance, the same benchmark design can be recalibrated for other privacy-sensitive, self-hosted services.

Larger fleets may smooth transient imbalance through additional replicas, but they may also reproduce the same bottleneck when strong-tier demand grows faster than strong-tier capacity. We therefore present the released configurations as a factorized benchmark template, not a universal fleet recipe. Applying it to another environment requires recalibrating request rates and routing thresholds and, where decisions depend on small differences, repeating runs to quantify uncertainty.

\subsection{Limitations}

The main limitation is the measurement scope. Our MIMIC-BHC analysis uses overlap- and coverage-oriented proxies, which compare long-context content preservation but do not measure hallucinations, contradictions, omission severity, clinical safety, or clinician-assessed usefulness. Automatic metrics only partially track physician judgment \cite{zhang2025llmevalmed,rauf2026practice}; consequently, our quality results support relative coverage trends, not clinical factuality or safety claims. No clinician review was performed. Before concluding patient-care suitability, we would need to evaluate factual correctness, omission severity, and workflow-specific latency targets.

The study uses one model family (Qwen2.5-7B/14B), one four-A100 hardware class across two execution environments, and one realized benchmark-backed request mixture. It is a controlled serving experiment rather than a replay of hospital or agent telemetry. It does not model retries, production bursts, urgency asymmetry, multi-call agent traces, or trace-driven context-length variation. Agent-driven demand is therefore motivation, not an evaluated workload. We also do not evaluate privacy controls, security, governance, or EHR integration; the clinical deployment context defines the capacity-planning scenario, not a validated hospital system. PubHealth and BioSSES are public benchmark resources; MIMIC-BHC is de-identified and used under its established access conditions. The serving experiments introduce no additional patient-identifying inputs. We do not measure energy, throughput per watt, GPU-hour cost, or cost per unit of quality. A fixed accelerator-slot budget is therefore an experimental resource constraint, not a demonstrated monetary or energy optimum.

The complete 24-cell matrix and the unrepeated follow-up policies remain single-run measurements; the reliability block covers ten decision-critical cells rather than every matrix-policy combination. Accordingly, millisecond-scale differences involving unrepeated policies should not be treated as statistically established. The cross-pod shift in absolute MIMIC overlap scores also shows that deterministic decoding does not eliminate environment-sensitive generation differences. The routing family is intentionally portable and interpretable; it does not cover learned routers, admission control, batching-aware schedulers, prefill/decode disaggregation, or live queue-aware control. Finally, we report the focused capacity-aware follow-up only for \threesonel and do not claim that its exact gains transfer to balanced fleets or larger clusters.

\section{Conclusion}

We studied capacity planning for resource-constrained clinical LLM serving, using mixed inference to expose heterogeneous request costs and measured coverage needs. Across the controlled four-GPU matrix, fleet composition produced larger observed shifts in TTFT, latency, throughput, GPU saturation, and long-context coverage than routing-policy swaps. All-small fleets provided the strongest responsiveness, mixed-model fleets recovered most measured long-context coverage, and all-large fleets served as coverage-heavy but less responsive reference points. A focused follow-up and 42-run reliability block show repeat-supported serving gains from capacity-aware gating in a fixed \threesonel fleet, together with a small measured coverage trade-off.
Within the evaluated regime, these results support selecting fleet composition before refining the tested routing rules. The contribution is a reproducible empirical characterization of that interaction, with serving gains and measured coverage trade-offs. Generalization to other model families, larger fleets, hospital traces, and clinical service-level objectives requires further evaluation.

\section*{Acknowledgment}
OpenAI ChatGPT assisted with drafting and editing clarifications of experimental scope and limitations in Sections~I and III--V and the captions of Figures~1 and~3. This assistance did not generate new experimental measurements.

\bibliographystyle{IEEEtran}
\IEEEtriggeratref{20}
\bibliography{custom}

\end{document}